# Rumor Detection on Twitter Using Multiloss Hierarchical BiLSTM with an Attenuation Factor

Yudianto Sujana[*], Jiawen Li[*], Hung-Yu Kao
Intelligent Knowledge Management Lab
Department of Computer Science and Information Engineering
National Cheng Kung University
Tainan, Taiwan
yudianto.sujana@staff.uns.ac.id, p78073012@gs.ncku.edu.tw,
hykao@mail.ncku.edu.tw

## Abstract

Social media platforms such as Twitter have become a breeding ground for unverified information or rumors. These rumors can threaten people's health, endanger the economy, and affect the stability of a country. Many researchers have developed models to classify rumors using traditional machine learning or vanilla deep learning models. However, previous studies on rumor detection have achieved low precision and are time consuming. Inspired by the hierarchical model and multitask learning, a multiloss hierarchical BiLSTM model with an attenuation factor is proposed in this paper. The model is divided into two BiLSTM modules: post level and event level. By means of this hierarchical structure, the model can extract deep in-formation from limited quantities of text. Each module has a loss function that helps to learn bilateral features and reduce the training time. An attenuation fac-tor is added at the post level to increase the accuracy. The results on two rumor datasets demonstrate that our model achieves better performance than that of state-of-the-art machine learning and vanilla deep learning models.

## 1 Introduction

Currently, social media has a significant influence on people's daily lives. With social media, people can share information, speak freely and reproduce news online conveniently. Take Twitter as an example: over 500 million new tweets are sent every day, that is, nearly 5787 tweets per second (Cooper, 2019). However, unverified information, or rumors, is also diffused in social media; therefore, in the absence of a rumor detection

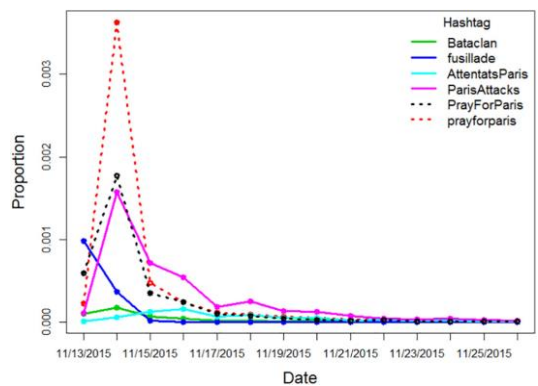

Figure 1: Relationship between the popularity of relevant hashtags about the Paris attack and time (Cvetojevic and Hochmair, 2018).

system, social media platforms can become a breeding ground for rumors.

In 2013, the Associated Press's official Twitter account was hacked and sent out a rumor that the president of the US was injured in an attack. This rumor caused wide panic and resulted in a brief crash of the stock market, in which investors lost $136 billion in just two minutes (Ajao et al., 2018). This incident highlighted that misinformation can threaten people's lives. Therefore, an automatic rumor detection system is vital. Information popularity in social media has a short lifetime; it usually stays for a few days, which is called the explosive increase phase. For instance, in Figure 1, tweets related to the Paris attack only stayed popular for two days (Cvetojevic and Hochmair, 2018). Unfortunately, Vosoughi et al. (2018) confirmed that false information propagated faster and was longer lasting than true information. Their research shows that it took nearly six times longer for verified information to reach 1500 people than for a rumor. According to the study, early as possible detection is highly practical to minimize harmful effects before rumors enter into the

---

[*] Equal contributions.

explosive increase phase. However, early rumor detection is the most difficult component of overall rumor detection. The greatest challenge lies in the lack of information.

To solve rumor detection problems, we analyze comments on a post. Comments can help in self-correcting the information dissemination through opinions, guesses and evidence shared by users. Thus, readers can judge the authenticity of information (Zubiaga et al., 2018). When commenting on an unverified post, people were inclined to use an interrogative or rhetorical tone (Kim, 2014; Ma et al., 2018a). Furthermore, rumors could be detected via the route of information diffusion(Ma et al., 2018b). However, the number of comments on a post is sometimes too narrow to use in early rumor detection. Therefore, making full use of limited information for accurate judgment remains a formidable challenge.

One important question is how to use such information. Zubiaga et al. (2016, 2017) proposed a method to use this information as a context to determine whether a tweet constituted a rumor. In contrast, some scholars suggested that events could be utilized as the basic processing unit for rumor detection, such as the tree-structured recursive neural network (Ma et al., 2018b), hierarchical structure model (Guo et al., 2018), and multitask learning (Ma et al., 2018a). Generally, an event contains an original post and a series of replies. Most of the scholars mentioned above use a large number of replies (from hundreds to thousands) to assist in detection. However, we believe that a large number of comments is not in line with the goal of early rumor detection; therefore, only one original post and a few early replies are used in this paper. Considering that the performance and capacity of a single processing layer to fully extract the text information is poor, we assume that higher-level structural models will bring more benefits for detection.

We attempted to represent information through a hierarchical neural network by building a post-level module first and an event-level module based on it. Since post-based and event-based rumor detection are highly related tasks, the hierarchical structure model can easily learn the bilateral feature representation based on these two tasks. In contrast to the traditional hierarchical structure, the model is based on a bidirectional long short-term memory (BiLSTM) model with some improvements. By means of the concept of multitask learning, we established a hierarchical model with a multiloss function to shorten the model training time and added an attenuation factor to the post-level model to maintain its precision. With this structure, the model can alleviate the impact of the vanishing gradient problem to a certain extent. The experimental results show that our model outperforms current state-of-the-art models.

Our contributions to this topic are as follows: (1) To the best of our knowledge, this is the first time that a multiloss function model with an attenuation factor was used for rumor detection. The model successfully combines post-level and event-level information for rumor detection. (2) The results of an evaluation using actual data from Twitter show that our model achieved high accuracy with only a few posts.

## 2 Related Work

Current automatic rumor detection systems suffer from low accuracy (Zubiaga et al., 2018). Two main approaches are used to debunk misinformation: the traditional method and the artificial intelligence approach. The traditional method manually analyzes text using statistics to define the critical features before detection. Castillo et al. (2011) proposed a large number of features for rumor detection by analyzing user attributes, rumor diffusion routes and text. Some researchers introduce various sets of features from different perspectives (Liu et al., 2015; K. Wu et al., 2015; Yang et al., 2012). With the development of artificial intelligence, some scholars have attempted to recognize rumors using deep learning. Ma et al. (2016) and Rath et al. (2017) used the RNN model to learn the abstract expression of rumors. Guo et al. (2018) proposed a hierarchical social attention model by combining a deep learning model and feature engineering, which improved the precision of rumor detection.

Early detection is the most challenging part of rumor detection. Many attempts at early rumor detection have been made. Wang et al. (2017) analyzed prominent features of rumor propagation and proposed a probabilistic model. Kwon et al. (2017) found that user and linguistic features could be used as important indicators in rumor detection. In addition to traditional methods, machine learning and deep learning have been applied to early rumor detection. Wu et al. (2015) proposed a graph-kernel-based hybrid SVM classifier that could capture high-order propagation patterns. Zhao et al. (2015) developed a technique based on searching for enquiry phrases that yielded good performance. Zhang et al. (2018) proposed a heterogeneous network for early rumor detection that reached 61% precision. T. Chen et al. (2018) used an RNN network with soft-attention structures. L. Wu et al. (2017) built a neural network framework consisting of inferring rumor categories, selecting discriminative features, and learning a rumor classifier. Moreover, Nguyen et al. (2017) presented an approach that leveraged convolutional neural networks for learning the hidden representations of each tweet in combination with a time series.

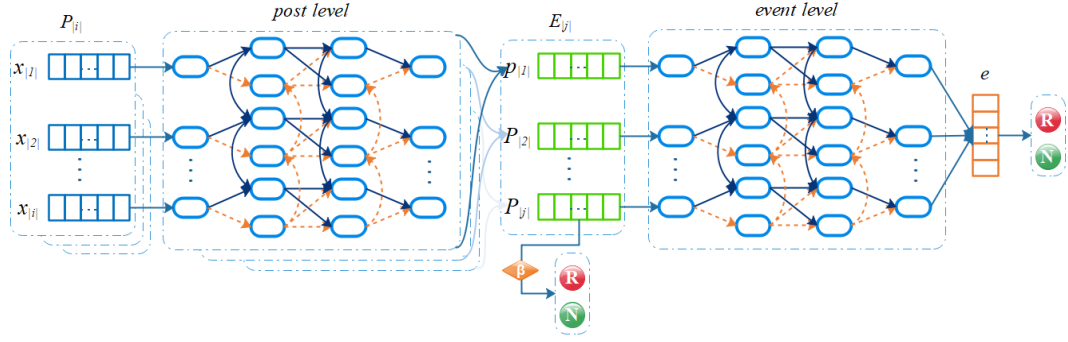

Figure 2: Multiloss BiLSTM hierarchical structure model with an attenuation factor.

Inspired by multitask learning and the hierarchical structure, we developed a multiloss hierarchical BiLSTM model with an attenuation factor that has high accuracy and performs well in early rumor detection.

## 3 Problem Formulation

A tweet consists of a limited number of words, some emojis and a few hashtags. This limited text makes it hard to classify misinformation. Therefore, we consider combining the source tweet and some of its comments as a whole event for rumor detection. We employ an event as the primary processing unit. An event contains more information and implicit users' stance (Liu et al., 2015; Lukasik et al., 2015, 2016; Mendoza et al., 2010; Rosengren et al., 2011).

Our hierarchical structure model begins with word embedding followed by post embedding and event embedding, with a fully connected layer at the end to detect whether the event is a rumor.

Multiple topics in the dataset are defined as $T = \{T_1, T_2, ... T_{|t|}\}$, and each $T_i$ contains multiple events, $T_i = \{E_1, E_2, ... E_{|e|}\}$. An event consists of a source post and a few comments, $E_e = \{P_s, P_1, P_3, ... P_{|p|}\}$. Notably, different topics have a different number of events, and events contain different numbers of posts, which means our model can handle variable length information. We develop this rumor detection task as a supervised classification problem. The classifier can perform learning via labeled information, that is, $f_e: \{E_1, E_2, ... E_{|e|}\} \to y_e$. At the same time, each post has its own label. Here, we define that each post label is identical to its corresponding source post and equivalent to the label of the event. Therefore, the post-level classifier $f_p: \{P_s, P_1, P_3, ... P_{|p|}\} \to y_p$ can be established, and all labels take one of two possible class labels: rumor or nonrumor.

## 4 Multiloss Hierarchical BiLSTM with an Attenuation Factor

The experimental results show that the hierarchical structure has strong information expression ability. However, our observations indicate that the hierarchical model has certain deficiencies, and backpropagation has to go through the time steps of all previous layers, which is computationally expensive and inefficient. It may also lead to vanishing gradient problems and substantially increase the training time. To shorten the training time and improve the training efficiency, we proposed a multiloss BiLSTM hierarchical structure model. Compared to the regular hierarchical model, this multiloss model is equivalent to a multitask learning model that can benefit bilaterally from the information features among multiple related tasks. Rumor detection at the post level and event level represent two branches under this theme, and the representations learned in the post level can be shared and used to reinforce the feature learning at the event level. Importantly, the backpropagation of the post level can help to alleviate the vanishing gradient problem in the early stage; thus, the model is stable and the training time is reduced.

Although the multiloss function can accelerate model training, the accuracy is slightly reduced because both post and event factors are considered in the classification process. Therefore, we added an attenuation factor to the post level to decrease the training time and maintain the high accuracy simultaneously.

Taking the text of all posts under an event as the input, we first perform word-embedding processing, where the processed word can be expressed as a fixed-length text vector. The formula is as follows:

$$x_t = E \Theta x_t \qquad (1)$$

where $x_t$ is the $t_{th}$ word in a post and E is a special word-embedding matrix. This step is omitted from the model diagram.

Next, all the vectors with the post as the unit pass through the post-level BiLSTM layer in proper order. Here, a deep BiLSTM structure is used. For each time point t, the formula is as follows:

$$h_{POST_i} = BiLSTM(x_i, h_{POST_{i-1}}) \qquad (2)$$

According to the physical meaning of LSTM, the cell state $h_t$ of the uppermost LSTM at the last time point is used as the result of the post encoding. Due to the use of the bidirectional structure, the final state of both directions is jointed, and an event can be represented by a matrix in which each column is a vector representing a post. The formula is as follows:

$$X = [h_{LSTM_{PS}}, h_{LSTM_{P1}}, h_{LSTM_{P2}}, \ldots, h_{LSTM_{P|p|}}] \qquad (3)$$

where $h_{LSTM_{Pi}}$ is the result from the post-level BiLSTM, that is, the embedding of one post.

The event-level BiLSTM formula is similar to the post-level BiLSTM. The difference is the input, where post-level BiLSTM uses $x_i$ and the event-level BiLSTM uses $X_I$:

$$h_{EVENT_i} = BiLSTM(X_I, h_{EVENT_{i-1}}) \qquad (4)$$

In the rumor detection classification task, the state of the event-level BiLSTM top layer at the last time point can be understood as an abstract representation of all post understandings.

To shorten the training time, the concept of multitask learning is used as a reference to realize the detection tasks of both posts and events. These two tasks are highly correlated, and the parameters in the post layer can be understood as common features. From the perspective of multitasking, the common feature region can assist the model training for the purpose of rapid convergence.

A post-level classifier and an event-level classifier are included in the model. Note that the post-level classifier functions only as an auxiliary convergence, so the post-level classifier classifies and backpropagates only the last set of posts for each event.

$$y_p = soft\,max(W_p * h_{LSTM_{p|c|}} + b_p)$$
$$y_e = soft\,max(W_e * h_{EVENT_{|c|}} + b_e) \qquad (5)$$

where $y_p$ and $y_e$ represent the post and event classification results, respectively, $W_p$ and $W_e$ are the weights of the fully connected layers, and $b_p$ and $b_e$ are the biases.

The multiloss function helps to achieve rapid convergence, but it reduces the accuracy. To realize the rapid training of the model while maintaining its precision, an attenuation factor is added in the backpropagation. Since the Adam optimizer is used, the formula is as follows:

$$g_e = \nabla_{\theta_{t-1}} f(\theta_{t-1})$$
$$m_t = \mu * m_{t-1} + (1 - \mu) * g_t$$
$$n_t = v * n_{t-1} + (1 - v) * g_t^2$$
$$\hat{m}_t = m_t / 1 - \mu^t \qquad (6)$$
$$\hat{n}_t = n_t / 1 - v^t$$
$$\Delta\theta_t^e = -\eta * \hat{m}_t^e / \sqrt{\hat{n}_t^e} + \varepsilon \qquad (7)$$
$$\Delta\theta_t^p = -\left(\eta * \hat{m}_t^p / \sqrt{\hat{n}_t^p} + \varepsilon\right) * \beta \qquad (8)$$

where $\hat{m}_t$ and $\hat{n}_t$ are the corrections of $m_t$ and $n_t$, respectively. $m_t$ and $n_t$ are the first-order and second-order moment estimates of the gradient under the event, respectively, which can be regarded as estimates of the expectation and be approximated as unbiased estimates. $[\mu, v, \varepsilon]$ are hyperparameters, and $\hat{m}_t^e$, $\hat{n}_t^e$, and $\theta_t^e$ represent the corresponding parameters of an event. β is an attenuation factor, which decreases to zero as the number of training epochs increases.

# 5 Experiments and Results

## 5.1 Data Collection

The data from two rumor datasets used in this study are derived from tweets posted during breaking news. Table 2 describes the statistics of these two datasets. Moreover, the two datasets contain a large number of properties that can be used for feature engineering, which is helpful for rumor detection. However, since we build a model based on deep learning, the model learns the features automatically from the posts.

| Statistic | PHEME 2017 | PHEME 2018 |
|---|---|---|
| Users | 49,345 | 50.593 |
| Posts | 103,212 | 105,354 |
| Events | 5,802 | 6,425 |
| Avg words/post | 13.6 | 13.6 |
| Avg posts/event | 17.8 | 16.3 |
| Max posts/event | 346 | 246 |
| Rumor | 1972 | 2402 |
| Nonrumor | 3830 | 4023 |
| Balance degree | 34.00% | 37.40% |

Table 2: Dataset statistics

## 5.2 Model Training

For our experiment, the datasets were randomly split: 80% for training, 10% for validation, and 10% for testing. Similar to the work of Ma et al. (2016), we calculated the accuracy, precision, recall and F1-score to measure the rumor detection performance.

In the data preprocessing phase, our data were subjected to the following processes: standardizing text and deleting useless network labels, emojis, etc. However, the stop words were retained because they contain words that can be used to reflect the emotions of the writer. We trained all the models by employing the derivative of the loss function through backpropagation and used the Adam optimizer to update the parameters. For the hyperparameters, the maximum value of vocabulary is 25000, the batch size is 64, the dropout rate is 0.5, the hidden size unit is 256, and the learning rate is 0.0001. Training was then performed based on different events until the loss value converged or the maximum number of epochs was reached.

| Dataset | Method | Acc | Pre | Rec | F1 |
|---|---|---|---|---|---|
| PHEME 2017 | SVM-BOW | 0.669 | 0.535 | 0.524 | 0.529 |
| | CNN | 0.787 | 0.737 | 0.702 | 0.719 |
| | BiLSTM | 0.795 | 0.763 | 0.691 | 0.725 |
| | BERT | 0.865 | **0.859** | 0.851 | 0.855 |
| | RDM* | 0.873 | 0.817 | 0.823 | 0.820 |
| | **MHA** | **0.926** | 0.834 | **0.956** | **0.891** |
| PHEME 2018 | SVM-BOW | 0.688 | 0.518 | 0.512 | 0.515 |
| | CNN | 0.795 | 0.731 | 0.673 | 0.701 |
| | BiLSTM | 0.794 | 0.727 | 0.677 | 0.701 |
| | BERT | 0.844 | 0.834 | 0.835 | 0.834 |
| | RDM* | 0.858 | 0.847 | 0.859 | 0.853 |
| | **MHA** | **0.919** | **0.892** | **0.923** | **0.907** |

Table 1: Comparison results

## 5.3 Result

We compare our model with the following models:

- SVM-BOW: SVM classifier using bag-of-words and N-gram (e.g., 1-gram, bigram and trigram) features (Ma et al., 2018b).

- CNN: A convolutional neural network model (Y.-C. Chen et al., 2017) for obtaining the representation of each tweet and classifying tweets with a softmax layer.

- BiLSTM: A bidirectional RNN-based tweet model (Augenstein et al., 2016) that considers the bidirectional context between the target and tweet.

- BERT: A fine-tuned BERT to detect rumors.

- RDM: A method that integrates GRU and reinforcement learning to detect rumors in the early stage (Zhou et al., 2019).

- MHA: Our hierarchical model with a multiloss function and an attenuation factor.

The results of all the methods are illustrated in Table 1, and the MHA model yields the best performance. The SVM-BOW result is poor because the traditional statistical machine learning method is not able to capture helpful features in this complicated rumor detection task. For the CNN, BiLSTM, and RDM models, the results are worse than those of our model due to the insufficient capacity for information extraction. Those models process information based on posts and cannot obtain high-level representations from a hierarchical structure. BERT achieves state-of-the-art performance in many other NLP tasks. It has multiple layers and multihead attention and can mine in-depth information, but this structure is also based on posts and does not consider the post-event structure.

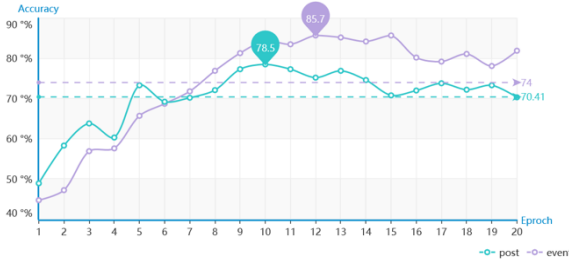

Figure 4: Comparison between post and event-based detection.

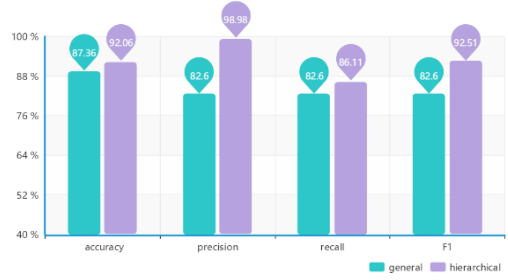

Figure 5: Comparison between the general BiLSTM model and hierarchical BiLSTM.

### 5.4 Ablation Experiments

**Event and Post Analysis.** We suggest that rumor detection based on an event is more credible than rumor detection based on a post. To prove this point, we conducted an experiment in which two models with identical structures and parameters were used to detect rumors in two different datasets. These two datasets contain the same text information: the only difference is that one is based on posts and the other is based on events. From the experimental results shown in Figure 4, we can see that the accuracy of the model with the event dataset is approximately 7% higher. This result verifies our assumption that rumor detection with events as the detecting unit is more accurate. Meanwhile, such an idea also paves the way for us to develop the hierarchical structure.

**General Structure and Hierarchical Structure Comparison.** We believe that the hierarchical structure, which has an advanced processing unit, is superior to the general structure in terms of extracting more complex and more in-depth information. To prove our hypothesis, we compared the general BiLSTM with the hierarchical BiLSTM (post-event layer). The hierarchical structure has two levels, namely, the post and event, in which the output from the post level becomes the input of the event level.

We used the same parameters for each module to ensure a fair comparison. Figure 5 shows that the hierarchical structure outperforms the general structure in terms of accuracy, precision, recall, and F1, which confirms our hypothesis that the hierarchical structure has stronger detection capability.

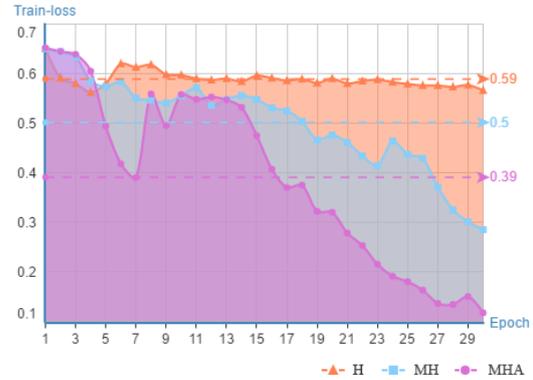

Figure 3: Loss comparison.

**Effects of Multiloss Functions and Attenuation Factor.** We also evaluated several internal models to show how the multiloss function helps in rumor detection and to further investigate the impact of the attenuation factor in the proposed model:

- Hierarchical (H): BiLSTM hierarchical structure model.

- Multiloss Hierarchical (MH): Multiloss BiLSTM hierarchical structure model.

- MHA: Multiloss BiLSTM hierarchical structure model with an attenuation factor

We compared the training results of the H, MH, and MHA models on the same dataset with the same random seed. Figure 3 shows that the MH and MHA, which are multiloss function models, learn faster than the original hierarchy model in the first 30 epochs. Moreover, the loss in that model decreases sharply. This result proves that the models benefit from post-level backpropagation by applying a multiloss function.

| Test dataset | Number of posts in each event |
|---|---|
| Test_5 | 2 – 5 |
| Test_10 | 6 – 10 |
| Test_15 | 11 – 15 |
| Test_20 | 16 – 20 |
| Test_25 | 21 – 25 |
| Test_30 | 26 – 30 |

Table 3: Test dataset for early rumor detection

The attenuation factor in MHA gradually decreased to zero until epoch fifteen. This attenuation factor makes the MHA model learn based on only the event label, whereas the HA model continues to tune the parameters based on both post and event information. With this technique, the training process becomes faster while maintaining the loss decreases.

## 5.5 Early Rumor Detection

To evaluate the model's early rumor detection performance, we considered six types of test sets that reflect the real scenario of rumor spreading on Twitter.

A small number of posts for each event means that the rumor had just begun to spread, with only a few tweets about the rumor. On the other hand, a large number of tweets implies that the rumors have spread widely.

The test results shown in Figure 6 indicate that our MHA model detects rumors better and with higher accuracy in the Test_5 dataset than do the other methods. This result implies that our models can classify rumors very early. Furthermore, our model also performs well in other test datasets, which indicates that our model can be used to detect both new rumors and widely spread rumors.

## 6 Conclusion and Future Work

In this paper, we introduced a multiloss hierarchical BiLSTM with an attenuation factor model for rumor detection. By means of the hierarchical structure, the model can learn deeply from limited text. The multiloss function makes the model learn efficiently and robustly, while the attenuation factor at the post level helps to increase the accuracy of rumor detection. The experimental results based on two PHEME datasets demonstrate that the model consistently outperforms other models by a significant margin. The model represents any post and event text with a fixed size length vector, which means it has strong applicability for both early and widely spread rumor detection with only a few modifications. In the future, the model can be extended by implementing social feature engineering to analyze and track rumors.

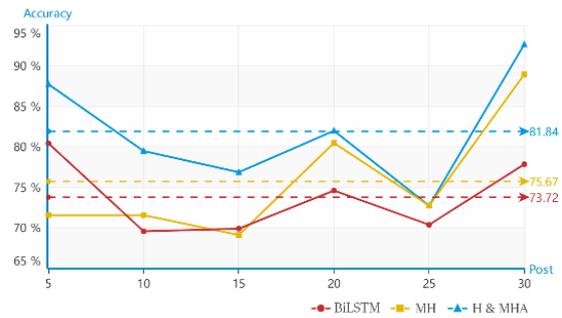

Figure 6: Early rumor detection accuracy at different numbers of posts.